\documentclass{article}
\usepackage{ijcai15}

\usepackage{graphicx}
\usepackage{algorithm}
\usepackage{algorithmic}
\usepackage{amsthm}
\usepackage{amssymb}
\usepackage{multicol}
\usepackage{amsmath}

\usepackage{bbm,amssymb,amsmath,amsfonts,graphicx,fullpage,ifthen,twoopt,algorithm,algorithmic,color}

\usepackage{times}

\title{A Survey on Practical Applications of Multi-Armed  and Contextual Bandits}
\author{Djallel Bouneffouf$^1$, Irina Rish$^2$\\
$^{1,2}$IBM Thomas J. Watson Research Center, Yorktown Heights, NY USA\\
\{dbouneffouf, Irish \}@us.ibm.com 
}

\begin{document}

\maketitle

\begin{abstract}
In recent years, multi-armed bandit (MAB) framework has attracted a lot of attention in various   applications, from   recommender systems and    information retrieval to healthcare   and  finance,   due to its stellar performance combined with certain attractive  properties, such as learning from less feedback.   The      multi-armed bandit field is currently flourishing, as  novel problem settings and algorithms  motivated by various practical applications  are being introduced, building on top of the classical bandit problem. This article aims to provide a comprehensive review of top recent developments in multiple real-life applications of the multi-armed bandit. Specifically, we introduce  a taxonomy of common MAB-based applications and summarize state-of-art for each of those domains. Furthermore, we identify important  current trends   and provide new perspectives pertaining to the future of  this exciting and fast-growing  field.
\end{abstract}

\section{Introduction}
Sequential decision-making problems, where at each   point in time an agent must choose the best action out of several alternatives, are frequently encountered in  various practical applications,   from clinical trials \cite{durand2018contextual} to recommender systems  \cite{MaryGP15} and anomaly detection   \cite{Ding:2019}.  Often, there is a side information, or  {\em context}, associated with each action (e.g., a user's profile), and the feedback, or {\em reward}, is limited to the chosen option. For example, in clinical trials \cite{durand2018contextual,bastani2015online}  the context is the patient's medical record (e.g. health condition, family history, etc.), the actions correspond to the treatment options being compared, and the reward represents the outcome  of the proposed treatment (e.g., success or failure). An important aspect affecting the long-term success in such settings is finding   a good trade-off between {\em exploration} (e.g., trying a new drug) and {\em exploitation} (choosing the  known drug).

This inherent exploration vs. exploitation trade-off exists in many sequential decision problems, and is traditionally formulated  as the {\em multi-armed bandit (MAB)} problem, stated as follows:   given $K$  possible actions, or ``arms'',  each associated with a fixed but unknown   reward probability distribution ~\cite {LR85,UCB},  at each iteration (time point) an agent selects   an arm to play   and receives a reward, sampled from the respective arm's probability distribution independently from the previous actions. The task of an agent is to learn how to choose its actions so that the  cumulative rewards over time is maximized. Different solutions have been proposed for this problem,  based on a stochastic formulation ~\cite {LR85,UCB,BouneffoufF16} and a Bayesian formulation ~\cite {AgrawalG12}; however, these approaches did not take into account the context, or side information available to the agent. 

A particularly useful version of the MAB is the {\em contextual multi-arm bandit (CMAB)}, or simply  the {\em contextual bandit} problem, where at  each iteration,  before choosing an arm, the agent observes an $N$-dimensional {\em context}, or {\em feature vector}.
The agent uses this context, along with the rewards of the arms played in the past, to choose which arm to play in the current iteration. Over time, the agent's aim is to collect enough information about the relationship between the context vectors and rewards, so that it can predict the next best arm to play by looking at the current context \cite{langford2008epoch,AgrawalG13}. Different algorithms were proposed for the general case, including LINUCB ~\cite{Li2010}, Neural Bandit \cite{AllesiardoFB14} and  Contextual Thompson Sampling (CTS)~\cite{AgrawalG13}, where a linear dependency is typically assumed between the expected reward of an action and its context. 

We will now provide an extensive overview including   various  applications of the bandit framework, both in real-life problem setting arising in multiple practical domains  (healthcare, computer network routing,  finance, and beyond),  as well as in computer science and  machine-learning in particular, where    bandit approaches can help   improve   hyperparameter tuning and other important algorithmic choices in supervised learning,   active learning and reinforcement learning. 

\section{Real-Life Applications of Bandit }
As a general mathematical framework, the stochastic multi-armed bandit setting addresses the primary difficulty in sequential decision-making under uncertainty, namely, the exploration versus exploitation dilemma, and therefore provides a natural formalism for most real-life  online decision making problems.
\subsection{Healthcare}

\textbf{Clinical trials. } Collecting data for assessing treatment effectiveness on animal models during the full range of disease stages can be difficult when using   conventional random treatment allocation procedures, since  poor treatments can cause deterioration of subject's health. Authors in \cite{durand2018contextual} aim to design an adaptive allocation strategy to improve the efficiency of data collection by allocating more samples for exploring promising treatments. They cast this application as a contextual bandit problem and introduce a practical algorithm for exploration vs. exploitation in this framework. The work relies on sub-sampling to compare treatment options using an equivalent amount of information. Precisely, they extend the sub-sampling strategy to contextual bandit setting g by applying sub-sampling within Gaussian Process regression. 

 Warfarin is the most widely used oral anticoagulant agent in the world; however,   dosing it correctly  remains a significant challenge, as the appropriate dose can be  highly variable among individuals due to various clinical, demographic and genetic factors. Physicians currently follow a fixed-dose strategy: they start patients on 5mg/day (the appropriate dose for the majority of patients) and slowly adjust the dose over the course of a few weeks by tracking the patient’s anti-coagulation levels. However, an incorrect initial dosage can result in highly adverse consequences such as stroke (if the initial dose is too low) or internal bleeding (if the initial dose is too high). Thus, authors in \cite{bastani2015online} tackle the problem of learning and assigning an appropriate initial dosage to patients by modeling the problem as a multi-armed bandit with high-dimensional covariates, and propose a novel and  efficient bandit algorithm based on the LASSO estimator.

\textbf{Brain and behavior modeling.} Drawing an inspiration from behavioral studies of human decision making in  both healthy controls and patients with different mental disorders, authors in \cite{bouneffouf2017bandit} propose a general parametric framework for multi-armed bandit problem which extends the standard Thompson Sampling approach to incorporate reward processing biases associated with several neurological and psychiatric conditions, including Parkinson's and Alzheimer's diseases, attention-deficit/hyperactivity disorder (ADHD), addiction, and chronic pain. They demonstrate empirically, from the behavioral modeling perspective, that their parametric framework can be viewed as a first step towards a unifying computational model capturing reward processing abnormalities across multiple mental conditions. 

\subsection{Finance}

In recent years, sequential portfolio selection has  been a focus of  increasing interest at the intersection of the machine learning and quantitative finance. The trade-off between exploration and exploitation, with the goal of  maximizing cumulative reward, is a natural formulation of the   portfolio choice problems. In \cite{shen2015portfolio},  the authors proposed   a bandit algorithm  for making online portfolio choices via exploiting correlations among multiple arms. By constructing orthogonal portfolios from multiple assets and integrating their approach with the upper-confidence-bound bandit framework, the authors derive the optimal portfolio strategy  representing  a combination of passive and active investments according to a risk-adjusted reward function. 
In \cite{huo2017risk}, the authors incorporate risk-awareness into the classic multi-armed bandit setting and introduce a novel algorithm for   portfolio construction. Through filtering assets based on the topological structure of financial market and combining the optimal multi-armed bandit policy with the minimization of a coherent risk measure, they achieve a balance between risk and return.

\subsection{Dynamic Pricing}
Online retailer companies are often faced with the dynamic pricing problem:  the company must decide on real-time prices for each of its multiple  products.  The company can run price experiments (make frequent price changes) to learn about demand and maximize long-run  profits. The authors in \cite{misra2018dynamic} propose a dynamic price experimentation policy, where the company has only  incomplete demand information. For  this  general  setting,  authors derive  a  pricing  algorithm  that  balances  earning  an immediate profit  vs. learning  for  future  profits.  The approach combines multi-armed  bandit   with partial identification of consumer demand from economic theory. Similar to \cite{misra2018dynamic}, authors in \cite{mueller2018low} consider high-dimensional dynamic multi-product pricing with an evolving low-dimensional linear demand model. They show that the revenue maximization problem reduces to an online bandit convex optimization with side information given by the observed demands. The approach applies a bandit convex optimization algorithm in a projected low-dimensional space spanned by the latent product features, while simultaneously learning this span via online singular value decomposition of a carefully-crafted matrix containing the observed demands.
 
\subsection{Recommender Systems}

Recommender systems are frequently used in various application  to predict user preferences. However, they also face  the exploration-exploitation dilemma when making a recommendation, since they need to  exploit their knowledge about the previously chosen items the user is  interested in, while also  exploring new items the user may like. Authors in \cite{zhou2017large} approach this challenge   using multi-armed bandit setting, especially for large-scale recommender systems that have really large or infinite number of items. They propose two large-scale bandit approaches in situations when no  prior information is available. Continuous exploration in their approaches can address the cold start problem in recommender systems. In context-aware recommender systems, most existing approaches focus on recommending relevant items to users, taking into account contextual information, such as time, location, or social aspects. However, none of those approaches has considered the problem of user’s content evolution. In \cite{bouneffouf2012contextual} authors introduce an algorithm that takes this dynamics into account. It is based on dynamic exploration/exploitation and can adaptively balance the two aspects, deciding which situation is most relevant for exploration or exploitation. 
In this sense, \cite{bouneffouf2014freshness} propose to study the "freshness" of the user's content through the bandit problem. They introduce in this paper an algorithm named Freshness-Aware Thompson Sampling that manages the recommendation of fresh document according to the user's risk of the situation.

\subsection{Influence Maximization}
Autors in \cite{vaswani2017model} consider influence maximization (IM) in social networks, which is the problem of maximizing the number of users that become aware of a product by selecting a set of “seed” users to expose the product to. They propose a novel parametrization that not only makes the framework agnostic to the underlying diffusion model, but also statistically efficient to learn from data. They give a corresponding monotone, submodular surrogate function, and show that it is a good approximation to the original IM objective. They also consider the case of a new marketer looking to exploit an existing social network, while simultaneously learning the factors governing information propagation. For this, they develop a LinUCB-based bandit algorithm. Authors in \cite{wen2017online} also study the online influence maximization problem in social networks but under the independent cascade model. Specifically, they try to learn the set of “best seeds or influencers” in a social network online while repeatedly interacting with it. They address the challenges of combinatorial action space, since the number of feasible influencer sets grows exponentially with the maximum number of influencers, and limited feedback, since only the influenced portion of the network is observed. They propose and analyze IMLinUCB, a computationally efficient UCB-based algorithm. 

\subsection{Information Retrieval}
Authors in \cite{losada2017multi} argue that Information Retrieval iterative selection process can be naturally modeled as a contextual bandit problem. Casting document judging as a multi-armed bandit problem leads to highly effective adjudication methods. Under this bandit allocation framework, they consider stationary and non-stationary models and propose seven new document adjudication methods (five stationary methods and two non-stationary variants). This comparative study includes existing methods designed for pooling-based evaluation and existing methods designed for metasearch. In mobile information retrieval, authors in \cite{bouneffouf2013contextual} introduce an algorithm that tackles this dilemma in Context-Based Information Retrieval (CBIR) area. It is based on dynamic exploration/exploitation and can adaptively balance the two aspects by deciding which user’s situation is most relevant for exploration or exploitation. Within a deliberately designed online framework they conduct evaluations with mobile users. 


\subsection{Dialogue Systems}
\textbf{Dialogue response selection.} Dialogue response selection is an important step towards natural response generation in conversational agents. Existing work on  conversational models mainly focuses on offline supervised learning using a large set of context-response pairs. In \cite{LiuYLM18} authors focus on online learning of response selection in dialog systems. They propose a contextual multi-armed bandit model with a nonlinear reward function that uses distributed representation of text for online response selection. A bidirectional LSTM is used to produce the distributed representations of dialog context and responses, which serve as the input to a contextual bandit. They propose a customized Thompson sampling method that is applied to a polynomial feature space in approximating the reward. 

\textbf{Pro-activity dialogue systems.} An objective of pro-activity in dialogue systems is to enhance the usability of conversational agents by enabling them to initiate conversation on their own. While dialogue systems have become increasingly popular during the last couple of years, current task-oriented dialogue systems are still mainly reactive, as users tend to initiate conversations. Authors of \cite{silander2018contextual} propose to introduce the paradigm of contextual bandits as framework for proactive dialog systems. Contextual bandits have been the model of choice for the problem of reward maximization with partial feedback since they fit well to the task description, they also explore the notion of memory into this paradigm, where they propose two differentiable memory models that act as parts of the parametric reward estimation function. The first one, Convolutional Selective Memory Networks, uses a selection of past interactions as part of the decision support. The second model, called Contextual Attentive Memory Network, implements a differentiable attention mechanism over the past interactions of the agent. The goal is to generalize the classic model of contextual bandits to settings where temporal information needs to be incorporated and leveraged in a learnable manner.
 
\textbf{Multi-domain dialogue systems.} Building multi-domain dialogue agents is a challenging task and an open problem in modern AI. Within the domain of dialogue, the ability to orchestrate multiple independently trained dialog agents, or skills, to create a unified system is of particular significance. In  \cite{upadhyaybandit}, the authors study the task of online posterior dialogue orchestration, where they define posterior orchestration as the task of selecting a subset of skills which most appropriately answers a user input using features extracted from both the user input and the individual skills. To account for the various costs associated with extracting skill features, they consider online posterior orchestration under a skill execution budget. This setting is  formalized as  Context-Attentive Bandit with Observations, a variant of context-attentive bandits, and evaluate it on simulated non-conversational and proprietary conversational datasets.

\subsection{Anomaly Detection}
 Performing anomaly detection on attributed networks concerns with finding nodes whose behaviors deviate significantly from the majority of nodes. Authors in \cite{Ding:2019} investigate the problem of anomaly detection in an interactive setting by allowing the system to proactively communicate with the human expert in making a limited number of queries about ground truth anomalies. Their objective is to maximize the true anomalies presented to the human expert after a given budget is used up. Along with this line, they formulate the problem through the principled multi-armed bandit framework and develop a novel collaborative contextual bandit algorithm, that explicitly models the nodal attributes and node dependencies seamlessly in a joint framework, and handles the exploration-exploitation dilemma when querying anomalies of different types. Credit card transactions predicted to be fraudulent by automated detection systems are typically handed over to human experts for verification. To limit costs, it is standard practice to select only the most suspicious transactions for investigation. Authors in \cite{soemers2018adapting} claim that a trade-off between exploration and exploitation is imperative to enable adaptation to changes in behavior. Exploration consists of the selection and investigation of transactions with the purpose of improving predictive models, and exploitation consists of investigating transactions detected to be suspicious. Modeling the detection of fraudulent transactions as rewarding, they use an incremental regression tree learner to create clusters of transactions with similar expected rewards. This enables the use of a {\em contextual} multi-armed bandit (CMAB) algorithm to provide the exploration/exploitation trade-off. 

\subsection{  Telecommunication}
In \cite{boldrini2018mumab}, a multi-armed bandit   model was used to describe the problem of best wireless network selection by a multi-Radio Access Technology (multi-RAT) device, with the goal of maximizing the quality perceived by the final user. The proposed model extends the classical MAB model in a twofold manner. First, it foresees two different actions: to measure and to use; second, it allows actions to span over multiple time steps. Two new algorithms designed to take advantage of the higher flexibility provided by the muMAB model are also introduced. The first one, referred to as measure-use-UCB1 is derived from the UCB1 algorithm, while the second one, referred to as Measure with Logarithmic Interval, is appositely designed for the new model so to take advantage of the new measure action, while aggressively using the best arm. 
The authors in \cite{KerkoucheAFVM18} demonstrate the possibility to optimize the performance of the Long Range Wide Area Network technology. Authors suggest that nodes use multi-armed bandit algorithms, to select the communication parameters (spreading factor and emission power). Evaluations show that such learning methods allow to manage the trade-off between energy consumption and packet loss much better than an Adaptive Data Rate algorithm adapting spreading factors and transmission powers on the basis of Signal to Interference and Noise Ratio values.

 \subsection{Bandit in Real-Life Applications: Summary and Future Directions}
 
\begin{table}[h]
\scriptsize
\caption {Bandit for Real Life Application}
\label{tab:Life} 
\begin{tabular}{|l|r|l|l|l|}
\hline
                            &     & Non-   &       & Non-   \\ 
                            & MAB     & stationary  &  CMAB      &  stationary  \\
                            &      & MAB    &       & CMAB  \\
                            \hline
Healthcare        &  $\surd$ &                      &  $\surd$  &                                  \\ \hline
Finance           &   $\surd$ &                     &           &                                  \\ \hline
Dynamic pricing   &         &     $\surd$           &           &                                  \\ \hline
Recommendr system & $\surd$  &    $\surd$           & $\surd$   &     $\surd$                          \\ \hline
Maximization      & $\surd$  &                      &           &                                  \\ \hline
Dialogue system   &          &                      &   $\surd$  &                                  \\ \hline
Telecomunication  &  $\surd$ &                      &           &                                  \\ \hline
Anomaly detection           &   $\surd$ &                      &           &                                  \\ \hline
\end{tabular}
\end{table}
Table \ref{tab:Life} provides a summary of bandit problem formulations used in each of the above  domain applications.  We see that, for example,  non-stationary bandit was not used in healthcare applications, since perhaps there was no significant change assumed to happen in the environment  in the process of making the treatment decision,  i.e. no  transition in the state of the  the patient;   such transition, if it occurred, would be better modeled using reinforcement learning rather than non-stationary bandit. There are clearly other domains where the non-stationary bandit is a more appropriate  setting, but it looks like   this setting was  not yet much investigated in healthcare domain. For example, anomaly detection, is a domain where non-stationary contextual bandit could be used, since in this setting the anomaly could be adversarial, which means that any bandit applied to this setting should have some kind of drift condition, in-order to adapt to new type of attack. Another observation  is that none of the existing work tried to develop an algorithm that could solve these different tasks at the same time, or apply the  knowledge obtained in one domain to another domain, thus opening a direction of research on {\em multitask} and {\em transfer learning} in bandit setting. Furthermore, given an online nature of bandit problem,  {\em continuous}, or {\em lifelong learning} would be a natural next step, adapting  the model learned in the previous tasks to the new one, while  still remembering how to perform earlier task, thus avoiding the  problem of ''catastrophic forgetting''.  

\section{Bandit for Better Machine Learning}
In this section we are describing how bandit algorithms could be used to improve  other algorithms, e.g. various machine-learning techniques.
\subsection{Algorithm Selection}
Algorithm selection is typically based on models of algorithm performance, learned during a separate offline training sequence, which can be prohibitively expensive. In recent work, they adopted an online approach, in which a performance model is iteratively updated and used to guide selection on a sequence of problem instances. The resulting exploration-exploitation trade-off was represented as a bandit problem with expert advice, using an existing solver for this game, but this required using an arbitrary bound on algorithm runtimes, thus invalidating the optimal regret of the solver. In  \cite{GaglioloS10}, a simpler framework was proposed for representing algorithm selection as a bandit problem, using partial information  and an unknown bound on losses. 

\subsection{Hyperparameter Optimization}
Performance of machine learning algorithms depends critically on identifying a good set of hyperparameters. While recent approaches use Bayesian optimization to adaptively select optimal hyperparameter configurations, they rather focus on speeding up random search through adaptive resource allocation and early-stopping. \cite{li2016hyperband} formulated hyperparameter optimization as a pure-exploration non-stochastic infinite-armed bandit problem where a predefined resources, such as iterations, data samples, or features are allocated to randomly sampled configurations. This work introduced a novel algorithm, Hyperband, for this framework and analyze its theoretical properties, providing several desirable guarantees. Furthermore,   Hyperband wascmpared with popular Bayesian optimization methods on a suite of hyperparameter optimization problems; it was observed that  Hyperband can provide more than an order-of-magnitude speedup over  its competitors on a variety of deep-learning and kernel-based learning problems.

\subsection{Feature Selection}
In a classical online {\em supervised learning}   the true  label of a sample is always revealed to the classifier, unlike in a bandit setting were any wrong classification resuls into zero reward, and only the single correct classification yields reward 1. The authors of~\cite{wang2014online} investigate the problem of Online Feature Selection, where the aim is to make accurate predictions using only a small number of active features using epsilon greedy algorithm. The authors of \cite{BouneffoufRCF17} tackle the online feature selection problem by addressing the combinatorial optimization problem in the stochastic bandit setting with bandit feedback, utilizing the Thompson Sampling algorithm. 

\subsection{Bandit for Active Learning}
Labelling all training examples in  supervised classification setting can be   costly. Active learning strategies solve this problem by selecting the most useful unlabelled examples to obtain the label for, and to train a predictive model. The choice of examples to label can be seen as a dilemma between the exploration and the exploitation over the input space. In \cite{bouneffouf2014contextual}, a novel active learning strategy manages this compromise by modelling the active learning problem as a contextual bandit problem.
they propose a sequential algorithm named Active Thompson Sampling (ATS), which, in each round, assigns a sampling distribution on the pool, samples one point from this distribution, and queries the oracle for this sample point label. The authors of  \cite{ganti2013building} also propose a multi-armed bandit inspired, pool-based active learning algorithm for the problem of binary classification. They utilize ideas such as lower confidence bounds, and self-concordant regularization from the multi-armed bandit literature to design their proposed algorithm.  In each round, the proposed algorithm assigns a sampling distribution on the pool, samples one point from this distribution, and queries the oracle for the label of this sampled point. 

\subsection{Clustering}
\cite{SublimeL18} considers collaborative clustering,  which is machine-learning paradigm concerned with the unsupervised analysis of complex multi-view data using several algorithms working together. Well-known applications of collaborative clustering include multiview clustering and distributed data clustering, where several algorithms exchange information in order to mutually improve each others. One of the key issue with multi-view and collaborative clustering is to assess which collaborations are going to be beneficial or detrimental. Many solutions have been proposed for this problem, and all of them conclude that, unless two models are very close, it is difficult to predict in advance the result of a collaboration. To address this problem, the authors of \cite{SublimeL18} propose a collaborative peer to peer clustering algorithm based on the principle of non stochastic multi-arm bandits to assess in real time which algorithms or views can bring useful information. 

\subsection{Reinforcement learning}
 Autonomous cyber-physical systems play a large role in our lives. To ensure that agents behave in ways aligned with the values of the societies in which they operate, we must develop techniques that allow these agents to not only maximize their reward in an environment, but also to learn and follow the implicit constraints assumed by the society. In \cite{noothigattu2018interpretable}, the authors study a setting where an agent can observe traces of behavior of members of the society but has no access to the explicit set of constraints that give rise to the observed behavior. Instead, inverse reinforcement learning is used to learn such constraints, that are then combined with a possibly orthogonal value function through the use of a contextual bandit-based orchestrator that picks a contextually-appropriate choice between the two policies (constraint-based and environment reward-based) when taking actions. The contextual bandit orchestrator allows the agent to mix policies in novel ways, taking the best actions from either a reward maximizing or constrained policy. The \cite{LarocheF17} tackles the problem of online RL algorithm selection. A meta-algorithm is given for input a portfolio constituted of several off-policy RL algorithms. It then determines at the beginning of each new trajectory, which algorithm in the portfolio is in control of the behaviour during the next trajectory, in order to maximise the return.  A novel meta-algorithm, called Epochal Stochastic Bandit Algorithm Selection. Its principle is to freeze the policy updates at each epoch, and to leave a rebooted stochastic bandit in charge of the algorithm selection. 
 
 \subsection{Bandit for Machine Learning: \\ Summary and Future Directions}
 
 \begin{table}[]
\scriptsize
\caption {Bandit in Machine Learning}
\label{tab:ML} 
\begin{tabular}{|l|r|l|l|l|}
\hline
                                  & MAB          & Non-         & CMAB        & Non-  \\ 
                                  &              & stationary  &         & stationary \\ 
                                  &              & MAB        &         & CMAB \\ 
                                  \hline
Algorithm Slection      &              & $\surd$                   &             &                                  \\ \hline
Parameter Optimization  &    $\surd$   &                           &            &                                  \\ \hline
Features Selection     &    $\surd$   &       $\surd$             &            &                                  \\ \hline
Active Learning        &    $\surd$   &                            &   $\surd$    &                                  \\ \hline
Clustering             &   $\surd$    &                           &            &                                  \\ \hline
Reinforcement learning  &  $\surd$     &       $\surd$                &    $\surd$         &                                  \\ \hline
\end{tabular}
\end{table}
 Table \ref{tab:ML} summarizes  the types of bandit problems   used to solve the machine-learning problems mentioned above. We see, for example, that contextual bandit was not used in feature selection or hyperparameter optimization. This   observation could point into a direction for future  work, where side information could be employed in feature selection. Also,     non-stationary bandit was   rarely considered in these problem settings, which is also suggesting possible extensions of current work. For instance, the non-stationary contextual bandit could be useful in the non-stationary feature selection setting, where finding the right features is time-dependent and context-dependent when the environment keeps changing. Our main observation   is also that each technique is solving just one machine learning problem at a time;  thus, the question is whether  a bandit setting and algoritms can be developed to solve multiple machine learning problems simultaneously, and whether transfer and continual learning can be achieved in this setting.  One solution  could be to model all these problems in a combinatorial bandit framework, where the bandit algorithm would find the optimal solution for each problem at each iteration; thus, combinatorial bandit could be further used as a tool for advancing  automated machine learning.

\section{Conclusions}
\label{sec:Conclusion}
In this article, we reviewed some of the most notable recent work   on applications of multi-armed bandit and contextual bandit, both in real-life domains and in automated machine learning. We summarized, in an organized way (Tables 1 and 2),  various  existing applications, by types of bandit settings used, and discussed the advantages of using bandit techniques in each domain. We briefly outlines of several important open problems and promising future extensions. 

In summary, the bandit framework, including both multi-arm and contextual bandit, is currently very active and promising research areas, and there are multiple novel techniques and applications emerging each year. We hope our  survey can help the reader better understand some key aspects of this exciting field and  get a better perspective on   its notable advancements and future promises.


\bibliographystyle{named}
\bibliography{ijcai15}

\begin{thebibliography}{}

\bibitem[\protect\citeauthoryear{Agrawal and Goyal}{2012}]{AgrawalG12}
Shipra Agrawal and Navin Goyal.
\newblock Analysis of thompson sampling for the multi-armed bandit problem.
\newblock In {\em {COLT} 2012 - The 25th Annual Conference on Learning Theory,
  June 25-27, 2012, Edinburgh, Scotland}, pages 39.1--39.26, 2012.

\bibitem[\protect\citeauthoryear{Agrawal and Goyal}{2013}]{AgrawalG13}
Shipra Agrawal and Navin Goyal.
\newblock Thompson sampling for contextual bandits with linear payoffs.
\newblock In {\em ICML (3)}, pages 127--135, 2013.

\bibitem[\protect\citeauthoryear{Allesiardo \bgroup \em et al.\egroup
  }{2014}]{AllesiardoFB14}
Robin Allesiardo, Rapha{\"{e}}l F{\'{e}}raud, and Djallel Bouneffouf.
\newblock A neural networks committee for the contextual bandit problem.
\newblock In {\em Neural Information Processing - 21st International
  Conference, {ICONIP} 2014, Kuching, Malaysia, November 3-6, 2014.
  Proceedings, Part {I}}, pages 374--381, 2014.

\bibitem[\protect\citeauthoryear{Auer \bgroup \em et al.\egroup }{2002}]{UCB}
Peter Auer, Nicol{\`o} Cesa-Bianchi, and Paul Fischer.
\newblock Finite-time analysis of the multiarmed bandit problem.
\newblock {\em Machine Learning}, 47(2-3):235--256, 2002.

\bibitem[\protect\citeauthoryear{Bastani and Bayati}{2015}]{bastani2015online}
Hamsa Bastani and Mohsen Bayati.
\newblock Online decision-making with high-dimensional covariates.
\newblock {\em Available at SSRN 2661896}, 2015.

\bibitem[\protect\citeauthoryear{Boldrini \bgroup \em et al.\egroup
  }{2018}]{boldrini2018mumab}
Stefano Boldrini, Luca De~Nardis, Giuseppe Caso, Mai Le, Jocelyn Fiorina, and
  Maria-Gabriella Di~Benedetto.
\newblock mumab: A multi-armed bandit model for wireless network selection.
\newblock {\em Algorithms}, 11(2):13, 2018.

\bibitem[\protect\citeauthoryear{Bouneffouf and
  F{\'{e}}raud}{2016}]{BouneffoufF16}
Djallel Bouneffouf and Rapha{\"{e}}l F{\'{e}}raud.
\newblock Multi-armed bandit problem with known trend.
\newblock {\em Neurocomputing}, 205:16--21, 2016.

\bibitem[\protect\citeauthoryear{Bouneffouf \bgroup \em et al.\egroup
  }{2012}]{bouneffouf2012contextual}
Djallel Bouneffouf, Amel Bouzeghoub, and Alda~Lopes Gan{\c{c}}arski.
\newblock A contextual-bandit algorithm for mobile context-aware recommender
  system.
\newblock In {\em International Conference on Neural Information Processing},
  pages 324--331. Springer, 2012.

\bibitem[\protect\citeauthoryear{Bouneffouf \bgroup \em et al.\egroup
  }{2013}]{bouneffouf2013contextual}
Djallel Bouneffouf, Amel Bouzeghoub, and Alda~Lopes Gan{\c{c}}arski.
\newblock Contextual bandits for context-based information retrieval.
\newblock In {\em International Conference on Neural Information Processing},
  pages 35--42. Springer, 2013.

\bibitem[\protect\citeauthoryear{Bouneffouf \bgroup \em et al.\egroup
  }{2014}]{bouneffouf2014contextual}
Djallel Bouneffouf, Romain Laroche, Tanguy Urvoy, Raphael F{\'e}raud, and Robin
  Allesiardo.
\newblock Contextual bandit for active learning: Active thompson sampling.
\newblock In {\em International Conference on Neural Information Processing},
  pages 405--412. Springer, 2014.

\bibitem[\protect\citeauthoryear{Bouneffouf \bgroup \em et al.\egroup
  }{2017a}]{bouneffouf2017bandit}
Djallel Bouneffouf, Irina Rish, and Guillermo~A Cecchi.
\newblock Bandit models of human behavior: Reward processing in mental
  disorders.
\newblock In {\em AGI}, pages 237--248. Springer, 2017.

\bibitem[\protect\citeauthoryear{Bouneffouf \bgroup \em et al.\egroup
  }{2017b}]{BouneffoufRCF17}
Djallel Bouneffouf, Irina Rish, Guillermo~A. Cecchi, and Rapha{\"{e}}l
  F{\'{e}}raud.
\newblock Context attentive bandits: Contextual bandit with restricted context.
\newblock In {\em IJCAI 2017, Melbourne, Australia, August 19-25, 2017}, pages
  1468--1475, 2017.

\bibitem[\protect\citeauthoryear{Bouneffouf}{2014}]{bouneffouf2014freshness}
Djallel Bouneffouf.
\newblock Freshness-aware thompson sampling.
\newblock In {\em International Conference on Neural Information Processing},
  pages 373--380. Springer, 2014.

\bibitem[\protect\citeauthoryear{Ding \bgroup \em et al.\egroup
  }{2019}]{Ding:2019}
Kaize Ding, Jundong Li, and Huan Liu.
\newblock Interactive anomaly detection on attributed networks.
\newblock In {\em Proceedings of the Twelfth ACM International Conference on
  Web Search and Data Mining}, WSDM '19, pages 357--365, New York, NY, USA,
  2019. ACM.

\bibitem[\protect\citeauthoryear{Durand \bgroup \em et al.\egroup
  }{2018}]{durand2018contextual}
Audrey Durand, Charis Achilleos, Demetris Iacovides, Katerina Strati,
  Georgios~D Mitsis, and Joelle Pineau.
\newblock Contextual bandits for adapting treatment in a mouse model of de novo
  carcinogenesis.
\newblock In {\em Machine Learning for Healthcare Conference}, pages 67--82,
  2018.

\bibitem[\protect\citeauthoryear{Gagliolo and Schmidhuber}{2010}]{GaglioloS10}
Matteo Gagliolo and J{\"{u}}rgen Schmidhuber.
\newblock Algorithm selection as a bandit problem with unbounded losses.
\newblock In {\em Learning and Intelligent Optimization, 4th International
  Conference, {LION} 4, Venice, Italy, January 18-22, 2010. Selected Papers},
  pages 82--96, 2010.

\bibitem[\protect\citeauthoryear{Ganti and Gray}{2013}]{ganti2013building}
Ravi Ganti and Alexander~G Gray.
\newblock Building bridges: Viewing active learning from the multi-armed bandit
  lens.
\newblock {\em arXiv preprint arXiv:1309.6830}, 2013.

\bibitem[\protect\citeauthoryear{Huo and Fu}{2017}]{huo2017risk}
Xiaoguang Huo and Feng Fu.
\newblock Risk-aware multi-armed bandit problem with application to portfolio
  selection.
\newblock {\em Royal Society open science}, 4(11):171377, 2017.

\bibitem[\protect\citeauthoryear{Kerkouche \bgroup \em et al.\egroup
  }{2018}]{KerkoucheAFVM18}
Raouf Kerkouche, R{\'{e}}da Alami, Rapha{\"{e}}l F{\'{e}}raud, Nad{\`{e}}ge
  Varsier, and Patrick Maill{\'{e}}.
\newblock Node-based optimization of lora transmissions with multi-armed bandit
  algorithms.
\newblock In {\em {ICT} 2018, Saint Malo, France, June 26-28, 2018}, pages
  521--526, 2018.

\bibitem[\protect\citeauthoryear{Lai and Robbins}{1985}]{LR85}
T.~L. Lai and Herbert Robbins.
\newblock Asymptotically efficient adaptive allocation rules.
\newblock {\em Advances in Applied Mathematics}, 6(1):4--22, 1985.

\bibitem[\protect\citeauthoryear{Langford and Zhang}{2008}]{langford2008epoch}
John Langford and Tong Zhang.
\newblock The epoch-greedy algorithm for multi-armed bandits with side
  information.
\newblock In {\em Advances in neural information processing systems}, pages
  817--824, 2008.

\bibitem[\protect\citeauthoryear{Laroche and F{\'{e}}raud}{2017}]{LarocheF17}
Romain Laroche and Rapha{\"{e}}l F{\'{e}}raud.
\newblock Algorithm selection of off-policy reinforcement learning algorithm.
\newblock {\em CoRR}, abs/1701.08810, 2017.

\bibitem[\protect\citeauthoryear{Li \bgroup \em et al.\egroup }{2010}]{Li2010}
Lihong Li, Wei Chu, John Langford, and Robert~E. Schapire.
\newblock A contextual-bandit approach to personalized news article
  recommendation.
\newblock {\em CoRR}, 2010.

\bibitem[\protect\citeauthoryear{Li \bgroup \em et al.\egroup
  }{2016}]{li2016hyperband}
Lisha Li, Kevin Jamieson, Giulia DeSalvo, Afshin Rostamizadeh, and Ameet
  Talwalkar.
\newblock Hyperband: A novel bandit-based approach to hyperparameter
  optimization.
\newblock {\em arXiv preprint arXiv:1603.06560}, 2016.

\bibitem[\protect\citeauthoryear{Liu \bgroup \em et al.\egroup
  }{2018}]{LiuYLM18}
Bing Liu, Tong Yu, Ian Lane, and Ole~J. Mengshoel.
\newblock Customized nonlinear bandits for online response selection in neural
  conversation models.
\newblock In {\em AAAI, 2018}, pages 5245--5252, 2018.

\bibitem[\protect\citeauthoryear{Losada \bgroup \em et al.\egroup
  }{2017}]{losada2017multi}
David~E Losada, Javier Parapar, and Alvaro Barreiro.
\newblock Multi-armed bandits for adjudicating documents in pooling-based
  evaluation of information retrieval systems.
\newblock {\em Information Processing \& Management}, 53(5):1005--1025, 2017.

\bibitem[\protect\citeauthoryear{Mary \bgroup \em et al.\egroup
  }{2015}]{MaryGP15}
J{\'{e}}r{\'{e}}mie Mary, Romaric Gaudel, and Philippe Preux.
\newblock Bandits and recommender systems.
\newblock In {\em Machine Learning, Optimization, and Big Data - First
  International Workshop, {MOD} 2015}, pages 325--336, 2015.

\bibitem[\protect\citeauthoryear{Misra \bgroup \em et al.\egroup
  }{2018}]{misra2018dynamic}
Kanishka Misra, Eric~M Schwartz, and Jacob Abernethy.
\newblock Dynamic online pricing with incomplete information using multi-armed
  bandit experiments.
\newblock 2018.

\bibitem[\protect\citeauthoryear{Mueller \bgroup \em et al.\egroup
  }{2018}]{mueller2018low}
Jonas Mueller, Vasilis Syrgkanis, and Matt Taddy.
\newblock Low-rank bandit methods for high-dimensional dynamic pricing.
\newblock {\em arXiv preprint arXiv:1801.10242}, 2018.

\bibitem[\protect\citeauthoryear{Noothigattu \bgroup \em et al.\egroup
  }{2018}]{noothigattu2018interpretable}
Ritesh Noothigattu, Djallel Bouneffouf, Nicholas Mattei, Rachita Chandra,
  Piyush Madan, Kush Varshney, Murray Campbell, Moninder Singh, and Francesca
  Rossi.
\newblock Interpretable multi-objective reinforcement learning through policy
  orchestration.
\newblock {\em arXiv preprint arXiv:1809.08343}, 2018.

\bibitem[\protect\citeauthoryear{Shen \bgroup \em et al.\egroup
  }{2015}]{shen2015portfolio}
Weiwei Shen, Jun Wang, Yu-Gang Jiang, and Hongyuan Zha.
\newblock Portfolio choices with orthogonal bandit learning.
\newblock In {\em Twenty-Fourth International Joint Conference on Artificial
  Intelligence}, 2015.

\bibitem[\protect\citeauthoryear{Silander and
  others}{2018}]{silander2018contextual}
Tomi Silander et~al.
\newblock Contextual memory bandit for pro-active dialog engagement.
\newblock 2018.

\bibitem[\protect\citeauthoryear{Soemers \bgroup \em et al.\egroup
  }{2018}]{soemers2018adapting}
Dennis~JNJ Soemers, Tim Brys, Kurt Driessens, Mark~HM Winands, and Ann
  Now{\'e}.
\newblock Adapting to concept drift in credit card transaction data streams
  using contextual bandits and decision trees.
\newblock In {\em AAAI}, 2018.

\bibitem[\protect\citeauthoryear{Sublime and Lefebvre}{2018}]{SublimeL18}
J{\'{e}}r{\'{e}}mie Sublime and Sylvain Lefebvre.
\newblock Collaborative clustering through constrained networks using bandit
  optimization.
\newblock In {\em 2018 International Joint Conference on Neural Networks,
  {IJCNN} 2018, Rio de Janeiro, Brazil, July 8-13, 2018}, pages 1--8, 2018.

\bibitem[\protect\citeauthoryear{Upadhyay \bgroup \em et al.\egroup
  }{2018}]{upadhyaybandit}
Sohini Upadhyay, Mayank Agarwal, Djallel Bounneffouf, and Yasaman Khazaeni.
\newblock A bandit approach to posterior dialog orchestration under a budget.
\newblock 2018.

\bibitem[\protect\citeauthoryear{Vaswani \bgroup \em et al.\egroup
  }{2017}]{vaswani2017model}
Sharan Vaswani, Branislav Kveton, Zheng Wen, Mohammad Ghavamzadeh, Laks~VS
  Lakshmanan, and Mark Schmidt.
\newblock Model-independent online learning for influence maximization.
\newblock In {\em Proceedings of the 34th International Conference on Machine
  Learning-Volume 70}, pages 3530--3539. JMLR. org, 2017.

\bibitem[\protect\citeauthoryear{Wang \bgroup \em et al.\egroup
  }{2014}]{wang2014online}
Jialei Wang, Peilin Zhao, Steven~CH Hoi, and Rong Jin.
\newblock Online feature selection and its applications.
\newblock {\em IEEE Transactions on Knowledge and Data Engineering},
  26(3):698--710, 2014.

\bibitem[\protect\citeauthoryear{Wen \bgroup \em et al.\egroup
  }{2017}]{wen2017online}
Zheng Wen, Branislav Kveton, Michal Valko, and Sharan Vaswani.
\newblock Online influence maximization under independent cascade model with
  semi-bandit feedback.
\newblock In {\em Advances in neural information processing systems}, pages
  3022--3032, 2017.

\bibitem[\protect\citeauthoryear{Zhou \bgroup \em et al.\egroup
  }{2017}]{zhou2017large}
Qian Zhou, XiaoFang Zhang, Jin Xu, and Bin Liang.
\newblock Large-scale bandit approaches for recommender systems.
\newblock In {\em International Conference on Neural Information Processing},
  pages 811--821. Springer, 2017.

\end{thebibliography}

\end{document}